\documentclass[conference]{IEEEtran}
\IEEEoverridecommandlockouts
\usepackage{cite}
\usepackage{amsmath,amssymb,amsfonts}
\usepackage{algorithmic}
\usepackage{graphicx}
\usepackage{textcomp}
\usepackage{xcolor}
\def\BibTeX{{\rm B\kern-.05em{\sc i\kern-.025em b}\kern-.08em
    T\kern-.1667em\lower.7ex\hbox{E}\kern-.125emX}}

\usepackage[textsize=footnotesize]{todonotes}

\usepackage{verbatim}
\usepackage[nolist]{acronym}
\usepackage{dirtytalk}
\usepackage{subcaption}
\usepackage{graphicx}
\usepackage{comment}
\usepackage{hyperref}
\usepackage{flushend}
\usepackage[font=small,labelfont=bf, skip=1pt]{caption}
\graphicspath{{./figures/}}
\usepackage{enumitem}

\usepackage{morefloats}
\usepackage{soul}

\newcommand{\change}[1]{\textcolor{black}{#1}}

\newacro{vr}[VR]{Virtual Reality}
\newacro{hmd}[HMD]{Head-Mounted Display}
\newacro{ems}[EMS]{Electrical Muscle Stimulation}
\newacro{imu}[IMU]{Inertial Measurement Unit}
\newacro{bci}[BCI]{Brain-Computer Interface}
\newacro{vwg}[VWG]{Virtual World Generator}
\newacro{moo}[MOO]{Multi-Objective Optimization}
\newacro{dof}[DoF]{Degree of Freedom}
\newacro{ssq}[SSQ]{Simulator Sickness Questionnaire}
\newacro{ddr}[DDR]{Differential Drive Robot}
\newacro{fov}[FOV]{Field-of-View}
\newacro{ar}[CR]{Coupled Rotations}
\newacro{ur}[UR]{Unwound Rotations}
\newacro{osf}[OSF]{Open Science Foundation}
\newacro{vims}[VIMS]{Visually Induced Motion Sickness}
\newacro{ros}[ROS]{Robot Operating System}

\begin{document}

\title{Unwinding Rotations Improves User Comfort \\ with Immersive Telepresence Robots

}

\author{Markku Suomalainen, Basak Sakcak, Adhi Widagdo, Juho Kalliokoski, Katherine J.~Mimnaugh, \\ Alexis P.~Chambers, Timo Ojala and Steven M.~LaValle \\ \textit{Center of Ubiquitous Computing, Faculty of Information Technology and Electrical Engineering}, \\ \textit{University of Oulu,} Oulu, Finland \\ firstname.surname@oulu.fi%
\thanks{This work was supported by Business Finland (project HUMOR 3656/31/2019); Academy of Finland (projects PERCEPT 322637, CHiMP 342556); and the European Research Council (project ILLUSIVE 101020977)}}
%
\IEEEaftertitletext{\vspace{-2\baselineskip}}

\maketitle

\begin{abstract}
We propose unwinding the rotations experienced by the user of an immersive telepresence robot to improve comfort and reduce VR sickness of the user. By immersive telepresence we refer to a situation where a 360\textdegree~camera on top of a mobile robot is streaming video and audio into a head-mounted display worn by a remote user possibly far away. Thus, it enables the user to be present at the robot's location, look around by turning the head and communicate with people near the robot. By unwinding the rotations of the camera frame, the user's viewpoint is not changed when the robot rotates. The user can change her viewpoint only by physically rotating in her local setting; as visual rotation without the corresponding vestibular stimulation is a major source of VR sickness, physical rotation by the user is expected to reduce VR sickness. We implemented unwinding the rotations for a simulated robot traversing a virtual environment and ran a user study (N=34) comparing unwinding rotations to user's viewpoint turning when the robot turns. Our results show that the users found unwound rotations more preferable and comfortable and that it reduced their level of VR sickness. We also present further results about the users' path integration capabilities, viewing directions, and subjective observations of the robot's speed and distances to simulated people and objects. 

\end{abstract}


\section{Introduction}\label{sec:intro}
\begin{figure}
\centering
\includegraphics[width=\columnwidth]{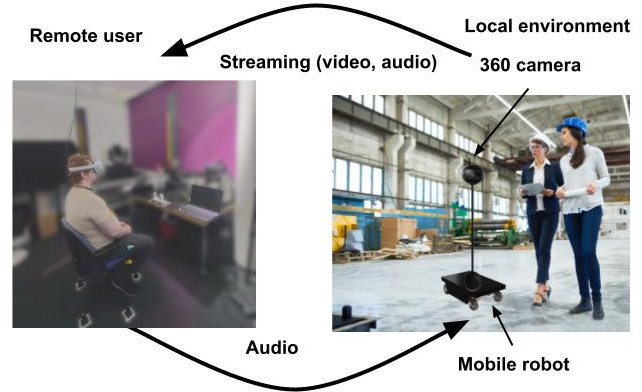}
\caption{Immersive robotic telepresence: a user wearing an \ac{hmd} can embody a robot far away and join events and facility tours even if physically joining is impossible. 
}
\label{fig:teleop} 
\vspace{-0.5cm}
\end{figure}

Immersive robotic telepresence refers to embodying a mobile robot in a far away location through a \ac{hmd}. The robot is equipped with either a 360\textdegree~camera or a regular camera with a fast pan-tilt-unit to follow the user's view, streaming video to the HMD such that he can look around in the robot's environment by turning his head (Fig.~\ref{fig:teleop}). This technology can greatly improve the experience for remote participants of \textit{hybrid} events where a group meets locally, but others who would like to join cannot be there physically. Use of an \ac{hmd} instead of a regular camera streaming into a regular screen as found in commercial telepresence robots (for example, GoBe and Double 3) has great potential to increase the \textit{presence} of the remote user, and thus making them feel as if they really were there at the robot's location. 

However, the use of an \ac{hmd} does not come without deficiencies, a major one being \ac{vr}
sickness \cite{laviola2000discussion,LaValle_bookVR} causing, for example, headache, blurred vision, and even vomiting. Different theories regarding the origin of VR sickness are all related to \textit{sensory mismatch} \cite{Reason75}; in immersive telepresence, a user's eyes observe motion but the vestibular organ does not since the user does not move. Moreover, the increased immersion leading to an increased feeling of presence can cause users to feel uncomfortable about other aspects besides sickness, such as distance to walls or surprising turns \cite{mimnaugh2021analysis,suomalainen2021comfort}. Thus, to make the benefits of immersive telepresence available for more people, care must be taken that the experience is comfortable and not sickening to users.

In this paper, we propose \textit{unwinding the rotations} for the user immersed in a telepresence robot. Essentially, this means that when the robot rotates, the user's viewpoint does not rotate along, such that the user always looks towards the same direction in the robot's environment unless they rotate themselves (assuming that the users are either standing or sitting in a swivelling chair). With the known result of rotational motion causing more sickness than translational motion \cite{kemeny2017,hu1999systematic}, we target the rotational motions in an attempt to reduce VR sickness. We predict that, according to the sensory mismatch hypothesis \cite{Reason75}, users will not get as sick when they have to perform the physical rotation themselves. 

The main contributions of this paper are 1) the method for unwinding rotations, compensating for the rotational motion that the camera undergoes due to the robot motions, and 2) a user study showing that unwinding rotations reduces VR sickness, increases user comfort and is preferred. \change{We performed the user study in simulation to eliminate possible confounding factors from a real robot causing additional sickness, such as vibrations or camera rotations around other axes.} We further show that users' path integration, the ability to mentally calculate updated position based on self-motions, does not deteriorate by unwinding the rotations. Additionally, we show that users rotate themselves in the chair to face the direction of motion, even though we did not prompt them to do it, and that they find it intuitive. Finally, we provide results from open-ended questions detailing reasons for preference of one method or the other, and update general recommendations regarding distances and speed of the robot in immersive telepresence. We believe that this paper makes an important contribution towards making immersive, HMD-based telepresence a more widely acceptable technology. 

\section{Related work}
\label{sec:related}
Telepresence, a term originally coined by Minsky \cite{minsky1980telepresence}, means being present in a remote location through technology.
To increase the mobility of the user, it is natural to mount a camera and a screen on a mobile robot, often referred to as a telepresence robot. There is evidence that even this kind of simple telepresence robots increase the feeling of \textit{presence}, the degree of ``being there" often measured in VR research \cite{skarbez2017survey}, of the remote user over a stationary video call \cite{rae2014bodies}. Research has shown multiple uses for robotic telepresence, such as students who otherwise cannot attend lectures \cite{rueben2021long,botev2021immersive},
social interaction \cite{kristoffersson2013review}, and telemonitoring on the basis of medical consultation \cite{carranza2018akibot}. 

Despite showing promise, the increase in presence from having wheels may still not be enough to 
enable communication comparable to physical presence: Stoll et al.~\cite{stoll2018wait} showed that users communicating through a robot spoke significantly less and perceived tasks and communication more difficult than local, physical participants. This was very likely related to the lack of presence, \change{since the feeling of presence is shown to make people act more naturally \cite{sanchez2005presence}}; despite the increase 
due to the wheels, the immersion level of a flat screen was not enough to create sufficient presence\footnote{We follow Mel Slater's framework for terminology, in which immersion is the physical system's capability of providing immersive experiences, and presence refers to the subjective feeling of being in a virtual location \cite{slater1997framework} }.
Combining the immersion of an \ac{hmd} with wheels, i.e., using an immersive telepresencce robot, has the potential to remedy this issue. Motivated by this potential, there has recently been an increasing amount of work about \ac{hmd}-based telepresence. The immersive capacity of \acp{hmd} in telepresence has been shown beneficial in, for example, education \cite{botev2021immersive}, general communication \cite{du2020human} and even a real earthquake scenario \cite{negrello2018humanoids}, and there are recent advances on controlling such a robot \cite{baker2020towards}. However, an important 
aspect still remains: how to make the robot's motions comfortable and non-sickening for the users. 

VR sickness is one of the major risks preventing widespread use of \acp{hmd}. \change{Whereas the umbrella term includes many issues causing users to feel sick, such as field of view and latency \cite{Chang_Kim_Yoo_2020}, in this paper we use} the term to refer to \ac{vims} in VR context. It is estimated that only 10\% of people have never experienced VIMS symptoms in a vehicle \cite{lawson2014motion}; as previous history in motion sickness is a good predictor for susceptibility to VR sickness \cite{golding1998motion}, it is easy to grasp the gravity of the problem. In mainstream VR research, the  problem is often alleviated by using teleportation or another locomotion model infeasible for a real robot. Beyond teleportation, other approaches to reduce VR sickness include slowing everything down (as optical flow correlates with experienced VR sickness \cite{laviola2000discussion}), using a rest frame, i.e., always keeping something constant in the user's view \cite{cao2018visually}, or narrowing the field of view \cite{Teixeira_Palmisano_2020}. However, all of these methods also deteriorate overall experience; slow motions are simply frustrating and a rest frame or narrower field of view reduce what the user sees and thus the feeling of presence \cite{weech2019presence}. 
Methods reducing VR sickness but not restricting the user's field of view are needed to make immersive telepresence more available for everyone.

Besides VR sickness and user comfort, another metric of interest in this study is the path integration of the user, which should not deteriorate when unwinding rotations, since disorientation can cause anxiety and discontent \cite{darken2002spatial}. Studies related to ours show that physical rotation increased spatial awareness (which correlates with path integration) over moving with a joystick \cite{riecke2010we} and a sviwel chair increases spatial awareness over a fixed chair \cite{hong2018effect}. 
In this study we do not explicitly tell the users to rotate themselves along with the robot while watching the videos, but they do sit in a swivel chair making it easily possible. 
We hypothesized there to be no difference in path integration between unwinding rotations and automatic, coupled rotations, a sufficient result if confirmed since reducing VR sickness is the main goal of unwinding rotations. We believe, however, that our study will be an interesting addition to the research in path integration as well.

\section{Unwinding Rotations}\label{sec:Unwinding}
\begin{figure}[t]
\centering
\includegraphics[width=0.94\columnwidth]{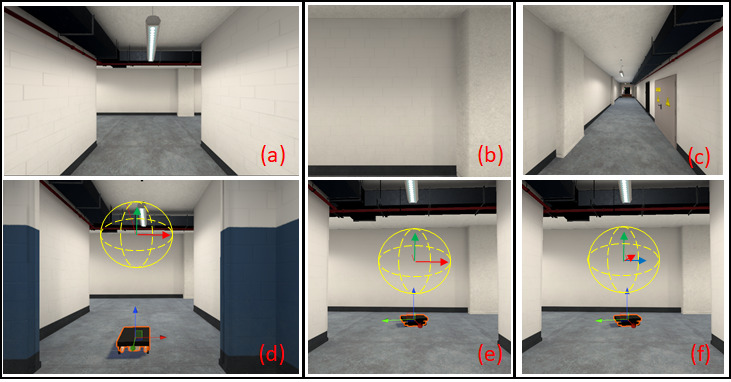}
\caption{The concept of unwinding rotations, assuming that user did not move her head. (a) Initial view of the immersed user. (b) View after the rotation is unwound. (c) View after the rotation is not unwound. (d) Initial robot configuration with robot and camera (in yellow sphere) coordinate frames. Robot and camera coordinate frames corresponding to (e) unwound and (f) not unwound rotation.  
}
\label{fig:unwinding} 
\vspace{-0.5cm}
\end{figure}

Unwinding rotations refers to rotating the camera frame to decouple the user's viewpoint from the rotations of the robot. 
This is illustrated in Fig.~\ref{fig:unwinding}: the robot is moving, either autonomously 
or controlled by a user directly, and will take a right at the corner.
Suppose the user's viewpoint before the turn is towards the wall at the end of the first corridor (view in Fig.~\ref{fig:unwinding}a). If we are not unwinding the 
rotation that the camera frame undergoes (Fig.~\ref{fig:unwinding}f) due to the robot motion, user's viewpoint will change towards the robot heading as the robot rotates, even if she has not moved her head (Fig.\ref{fig:unwinding}c). The proposed unwinding rotations method ensures that, by rotating the camera frame (Fig.~\ref{fig:unwinding}e), after the robot has made the turn, user's viewpoint will no longer be towards the robot's direction of motion but it will be towards the wall unless she physically moves her head (Fig.\ref{fig:unwinding}b). 

The main requirement for rotations to be unwound is that the camera mounted on the robot is capable of adjusting the user's viewpoint independently of the robot, through hardware (for example, a pan-tilt-unit) or software as in a 360\textdegree~camera. Let $\mathcal{R}_R$ be the set of all the rotations that can be applied to the robot to change its rotation, accordingly, let $\mathcal{R}^{-1}_R$ denote the set corresponding to the inverses of all the elements in $\mathcal{R}_R$. Note that, by the definition of the rotation matrix, the inverse always exists. Then, for unwinding rotations to be possible, the set of rotational transformations that can be applied to the camera frame $\mathcal{R}_C$ should contain $\mathcal{R}^{-1}_R$, i.e., $\mathcal{R}^{-1}_R \subseteq \mathcal{R}_C$. Essentially, this means that the camera frame should be capable of rotating to compensate for the robot rotation. For example, 
a 360\textdegree~camera can unwind any rotation through software, but a pan-tilt unit that cannot roll cannot be used to unwind rotations on a drone rotating about all three axes; however, such a pan-tilt-unit can be used to unwind all the rotations of a telepresence robot traversing a flat surface. Note that very small rotations can also be compensated using, for example, the post-rendering image warp \cite{mark1997post}. 


Consider a robot carrying a camera and moving in 3D space (for example, a drone or the end effector of a manipulator) such that $\mathcal{R}_R \in SO(3)$. Let $R_{R,k} \in \mathcal{R}_R$ be the rotation matrix corresponding to the robot orientation at time step $k \in \mathbb{Z}_{\geq0}$. We assume that the robot orientation, hence $R_{R,k}$, is known or can be estimated accurately at $k$, guaranteed for robotic systems which have sufficient sensing and filtering capabilities.
Then, we can define unwinding rotations as rotating the camera frame such that any point $p_c \in \mathbb{R}^3$ represented in the camera frame is related to the point $p'_c$, whose coordinates are expressed in the rotated camera frame, through
\begin{equation}
   p'_c=R_{C,k}p_c
   \label{eq:unwind_3D}
\end{equation}
in which $R_{C,k}=R_{R,k}^{-1} \in \mathcal{R}_C$. This operation negates the rotation of the viewpoint caused by the robot rotation. Thus, the users need to rotate themselves in order to face the direction that the robot is or face the same direction in the virtual environment for the whole motion.



The study presented in this paper considers a mobile robot moving in a two-dimensional plane, meaning that $R_{R,k}$ corresponding to the robot orientation $\theta_k$ at time step $k$ describes a rotation 
about the axis orthogonal to the plane that the robot is moving in. Therefore, for this special case, $\mathcal{R}_{R}$ is a subspace of $SO(3)$. 
In particular, if we consider that the axis about which the robot rotates is the $z$-axis (pointing up), then 
\begin{equation}
   R_{R,k}=\begin{bmatrix}
R(\theta_k) & 0_{2\times1}\\
0_{1\times2} & 1
\end{bmatrix}
\end{equation}
in which $R(\theta_k)\in SO(2)$ is the rotation in the $x\text{-}y$ plane. Then, we can use \eqref{eq:unwind_3D} to unwind the rotations applied to the camera frame due the changes in the robot orientation. 


\section{Hypotheses}\label{sec:hypotheses}
We pre-registered the following four hypotheses, together with the procedure and analyses to
be used in the study, in \ac{osf} \url{https://osf.io/eks6t}. From now on, we will refer to the \textbf{unwound rotations condition as UR}, for which the user's viewpoint does not change unless she physically rotates her head. Accordingly, we will refer to \textbf{\change{Coupled rotations as CR}}, for which the user's viewpoint always rotates when the robot rotates. 

\begin{enumerate}[label=\textbf{H\arabic*:}]
    \itemsep=0em
    \item 
    Less VR sickness in UR condition as indicated by lower total weighted \ac{ssq} score. 
    \item The UR condition is more comfortable as indicated by asking directly which was more comfortable (forced-choice). 
    \item The UR condition is preferred as indicated by asking directly which the user preferred (forced choice).
    \item No difference in path integration across conditions when measured by error in angle when asked to point towards the point of origin after the robot has finished the path.
\end{enumerate}

\textbf{H1} is based on the literature; as explained in Section~\ref{sec:related}, the sensory conflict theory postulates that a large part of VR sickness is caused by a mismatch between vision and vestibular organs. Therefore, we predicted that if the user experiences only the rotations that he performs physically, there will no sensory conflict; as a result, sickness should not be induced.
\textbf{H2} and \textbf{H3} were then prompted by the assumption that people do not want to be sick, and we did not foresee other major downsides that could make
UR not preferred or uncomfortable. \textbf{H4} is derived from the literature as well; there seemed to be
conflicting results regarding path integration, as presented in Section~\ref{sec:related}, which prompted us to propose no difference. The reason we test this
is to show that 
UR
does not deteriorate this important aspect of the experience.



\section{Experiments}\label{sec:methods}

\subsection{Study setup}
\label{sec:setup}
The image on the left in Fig.~\ref{fig:teleop} shows the setup used in this study. During the experiment, the participants sat on a swivel chair without wheels, to be able to easily rotate themselves as much as they wanted without translations possibly caused by wheels. We chose sitting over standing for a few reasons; first, many people seem to prefer sitting over standing in VR \cite{zielasko2021sit}; second, it is more inclusive since standing is physically more difficult for many people, and third; this way we can be sure that the people do not translate (take steps around), which could confound the study. To avoid the participants getting entangled with the cable, we installed a pulley system that feeds the cable from the ceiling.

\begin{figure*}%
  \centering
    \begin{subfigure}{0.43\columnwidth}
        \includegraphics[width=\columnwidth]{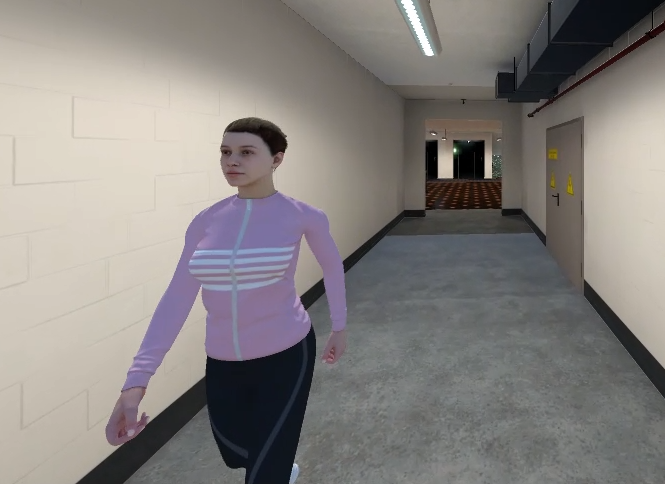}
         \caption{}
         \label{fig:person_walking}
    \end{subfigure}\hfill%
    \begin{subfigure}{0.55\columnwidth}
        \includegraphics[width=\columnwidth]{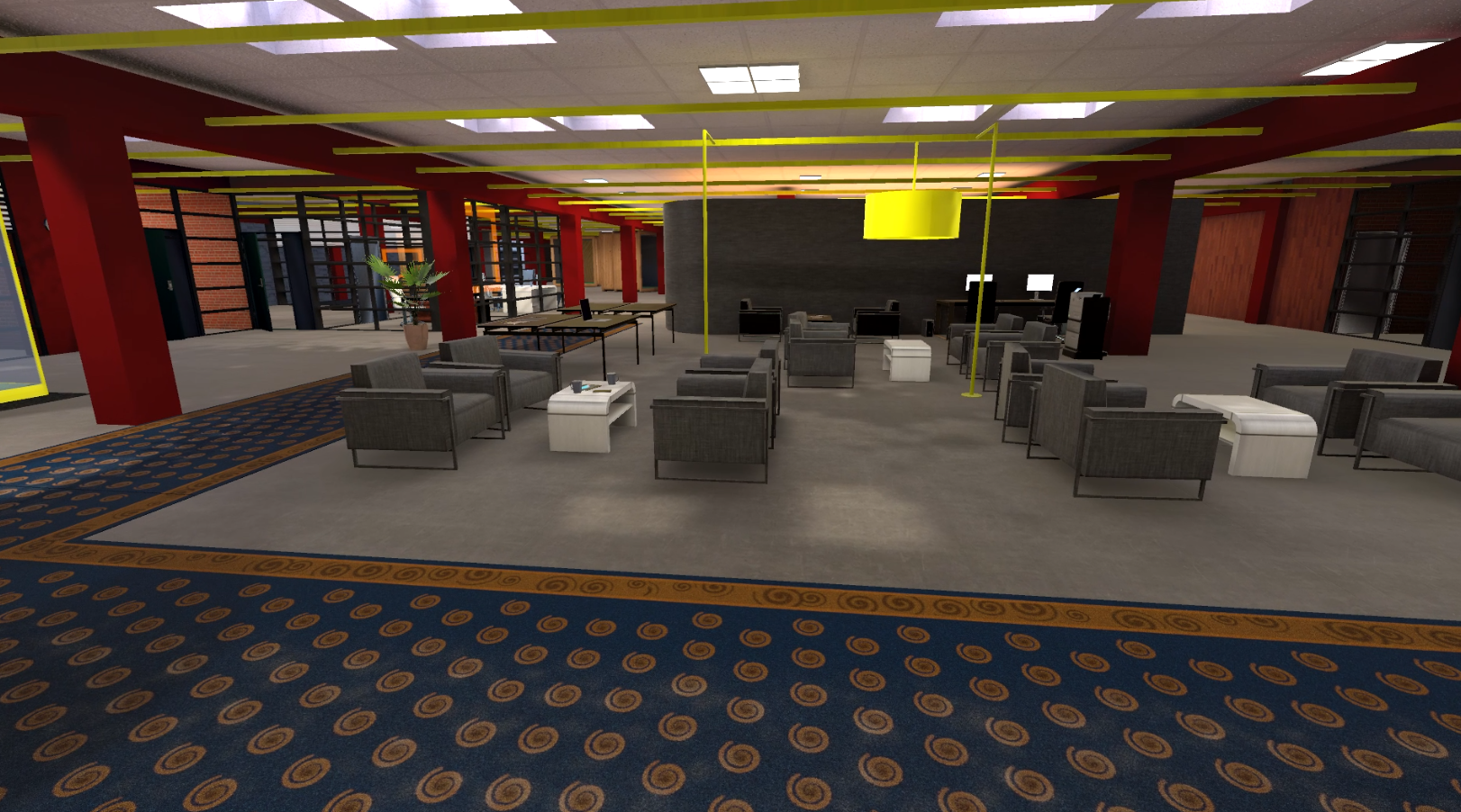}
        \caption{}\label{fig:general_view}
    \end{subfigure}\hfill%
    \begin{subfigure}{0.42\columnwidth}
        \includegraphics[width=\columnwidth]{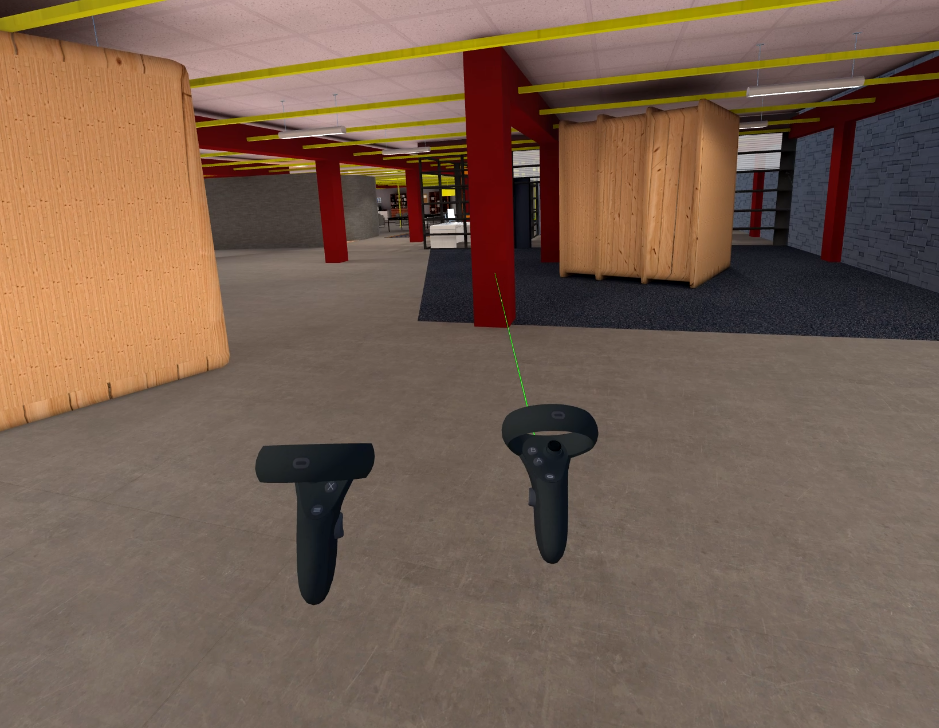}
         \caption{}\label{fig:pointing}
    \end{subfigure}\hfill%
    \begin{subfigure}{0.5\columnwidth}
       \includegraphics[width=\columnwidth]{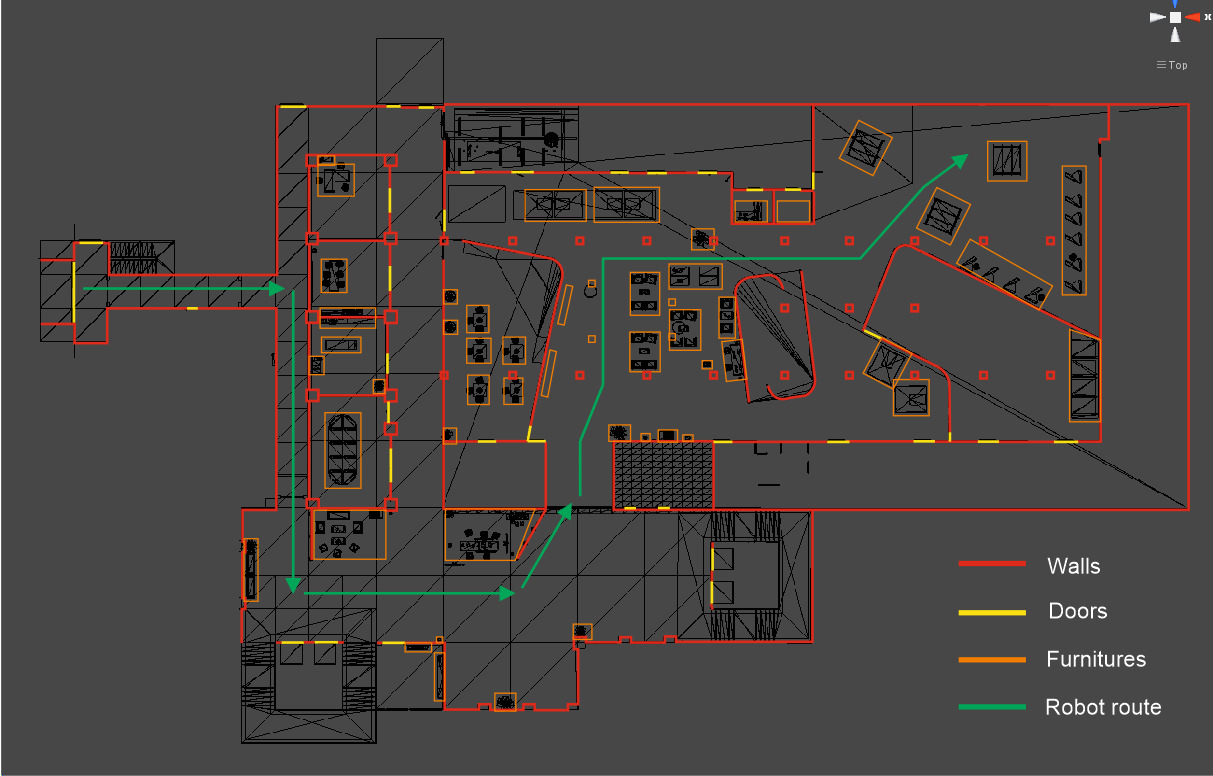}
    \caption{}\label{fig:map_path}
    \end{subfigure}%
\caption{Screenshots from the environment from the user's view; (a) a person passing by in a corridor; (b) general view; (c) raycasting with controllers at the end to point towards origin; (d) a birds-eye view of the environment with the taken path illustrated. }
\label{fig:screenshots}
\vspace{-0.5cm}
\end{figure*}

The virtual environment was designed using Blender and Unity. It is a combination of a real room at the University of Oulu and custom designed, imaginary areas (see Fig.~\ref{fig:screenshots} for screenshots from the environment); thus, even the people who are familiar with the university have no advantage in the path integration task. The virtual robot has been modelled after a real robot owned by the research group, a balanced differential drive robot with the 2 driving wheels in the middle and 4 caster wheels in corners to keep it steady (see Fig.~\ref{fig:unwinding}). We ensured that the dynamics governing the robot motion in simulation is sufficiently realistic. A virtual 360\textdegree~camera is attached 1.5 meters above the base, from which the user sees the virtual world; this is a height suggested by \cite{keskinen2019effect} for 360\textdegree~videos. To increase the ecological validity of the environment and the path, there are simulated people walking in the environment (Fig.~\ref{fig:person_walking}) that the robot occasionally evades. 

We used Nav2 project of \ac{ros} 2 Galactic for the navigation of the robot, and Unity as the physics-based simulator. The robot route is described as a sequence of waypoints, consequently, the path is calculated and tracked autonomously. In particular, we chose Theta* \cite{nash2010lazy} as the path planning algorithm and DWB (a controller based on Dynamic Window Approach \cite{fox1997dynamic}) as the controller to track the computed path. Since navigation is probabilistic in nature, due mainly to the localization and control methods used, one instance of the robot motion, together with the motion of the simulated humans, was recorded and then played back to the users in a Unity executable to avoid small differences in emerging motions. In the recorded motion, the robot moves at a maximum speed of $1\ m/s$ and rotates with velocities \change{within the range $[0,1]\ rad/s$ with maximum rotational acceleration $3.2\ rad/s^2$} as suggested in \cite{mimnaugh2021analysis,suomalainen2021comfort}. The path is $129.5$ meters long, the video lasts for 2 minutes 18 seconds, and the total amount of rotation that the robot executes during the video is $438^{\circ}$. The lowest distance to a wall or another inanimate object was $0.9m$ and the closest distance to people passing by $1\ m$; these values correspond to the distance accepted by users in \cite{mimnaugh2021analysis}. The application was viewed through 
Oculus Quest 2 via a link cable, with refresh rate 80Hz and resolution  5152x2608.

\subsection{Procedure}

The two videos of the UR and CR conditions were presented to the participants in a counterbalanced order such that both videos were seen first and second equal number of times by both men and women. Upon arrival, participants were greeted by a researcher and asked for consent, after which the experimenter asked the participants if they were feeling nauseous or had a headache in an effort to pre-screen people feeling sick already before the experiment (we would have re-scheduled a sick-feeling participant). Then, the users were asked to move to the swivel chair and adjust the chair height such that they could easily rotate around. Next, the experimenter read out the instructions, told the participant how to put on the HMD and asked them to rotate around once more while wearing the HMD; finally, the experimenter asked them to orient themselves towards ``the corridor with the fire hose", which was the direction the robot would start moving. Once this was confirmed and done (the experimenter could always see the participants' view on a mirrored screen) the experimenter started the video. During the video, the head orientation of the participant was recorded. 

When the video finished, the participants were asked not to take off the HMD but instead grab the controllers from a desk in front of them. Then, the users  were asked to point towards the beginning of the robot's path, which was clarified as the place with the fire hose to make sure participants understood this correctly. Then, the participants were asked to remove the HMD and fill out a questionnaire regarding their experience, after which the same procedure was repeated for the second video. At the end, the participants were rewarded with 20€ Amazon vouchers. Finally, we took the recommended precautions regarding Covid-19 when running the study. The experimenter was always wearing a masks and kept distance from the participant except if help was needed with the \ac{hmd}. With the \ac{hmd} disposable face covers were used. 

\subsection{Measures}
The path integration error was measured at the robot's final position as the absolute angle between the direction indicated by the participant and the direction of the origin (robot start position). Thus, the error lies within the interval $[0^\circ, 180^\circ]$. 
We computed the direction using the projection of the line segment protruding from the controller (see Fig.~\ref{fig:pointing}) onto the plane that the robot moves in. 

Head motion data was used to compute the average deviation from the robot heading to see whether participants aligned themselves with the direction of motion as they would if the rotations were not unwound. Deviation was calculated as the absolute angle difference between the robot orientation and the head orientation in the plane that the robot moves in; absolute angle was used to avoid bias due to the robot's turns not being equally divided between left and right. 
Furthermore, head orientation is plotted along the robot path to qualitatively analyze the emergent view patterns across conditions.  

In the questionnaires, we \change{measured sickness with the \ac{ssq} \cite{kennedy1993simulator}, an established questionnaire for measuring sickness in VR by presenting 16 possible sickness symptoms, which the participants gauge on a scale none (0) to severe (3). The answers are weighted for a maximum score of 236 \cite{bimberg2020usage}.} Additionally, we used 7-point Likert-scale questions, forced-choice questions comparing the two videos, and open-ended questions about reasons for some choices and demographic questions. The first question after each video was about the user's confidence in the path integration, after which SSQ was administered and we asked how comfortable the experience was on Likert-scale. After the second video, these questions were followed by forced-choice questions of choosing the preferred, more comfortable and more intuitive condition for the participant. Then, we asked Likert-scale questions about how the participants felt regarding distance to walls, distance to humans, and the linear speed of the robot; these questions were asked to make further suggestions for researchers working on autonomous motion planners for telepresence robots. Finally, we asked about VR and gaming experience and demographics. The whole questionnaire can be found in Appendix 1.

\subsection{Participants}
Based on the effect size from a power analysis of a previous study comparing two immersive telepresence experiences \cite{becerra2020human}, the necessary sample size was determined as 32 participants. We required an exact split between male and female participants due to earlier, contradicting evidence that gender may have an effect on the susceptibility of VR sickness, either women suffering more \cite{rebenitsch2014individual} or less \cite{Peck_Sockol_Hancock_2020} from VR sickness. When two participants from the first 32 recruited preferred not to report their gender, we recruited two more participants to be certain of the gender balance, ending up with 34 participants recruited from the university of Oulu campus and the surroundings. We used only the gender-specified 32 participants for testing for VR sickness differences between genders, and all 34 participants for all other analyses. Participants' age ranged from 18 to 48 with the mean at 26.21 ($sd=7.1$). All participants reported having normal or corrected-to-normal vision and none of the participants were colorblind. 
The responses of the participants to how often they use \ac{vr} systems were: $41.2\%$ never used any before, $35.3\%$ just a couple of times and at least once, $11.8\%$ once or twice a year, $3\%$ once or twice a month, and $8.8\%$ several times a week. Their responses to how often they play computer games were: $11.8\%$ never played before, $8.8\%$ once or just a couple of times ever, $20.6\%$ once or twice a year, $11.8\%$ once or twice a month, $17.6\%$ once or twice a week, $14.7\%$ several times a week, and $14.7\%$ every day. 


\section{Results}
Four confirmatory hypothesis tests were preregistered, as well as several topics for exploratory analysis. All tests were run in SPSS with (two-tailed) significance levels set to $0.05$ and with a $95 \%$ confidence interval. 

\subsection{Confirmatory Results}

\begin{figure}
     \centering
     \begin{subfigure}{0.49\columnwidth}
         \centering
         \includegraphics[trim=0 5 0 0,clip,width=\textwidth]{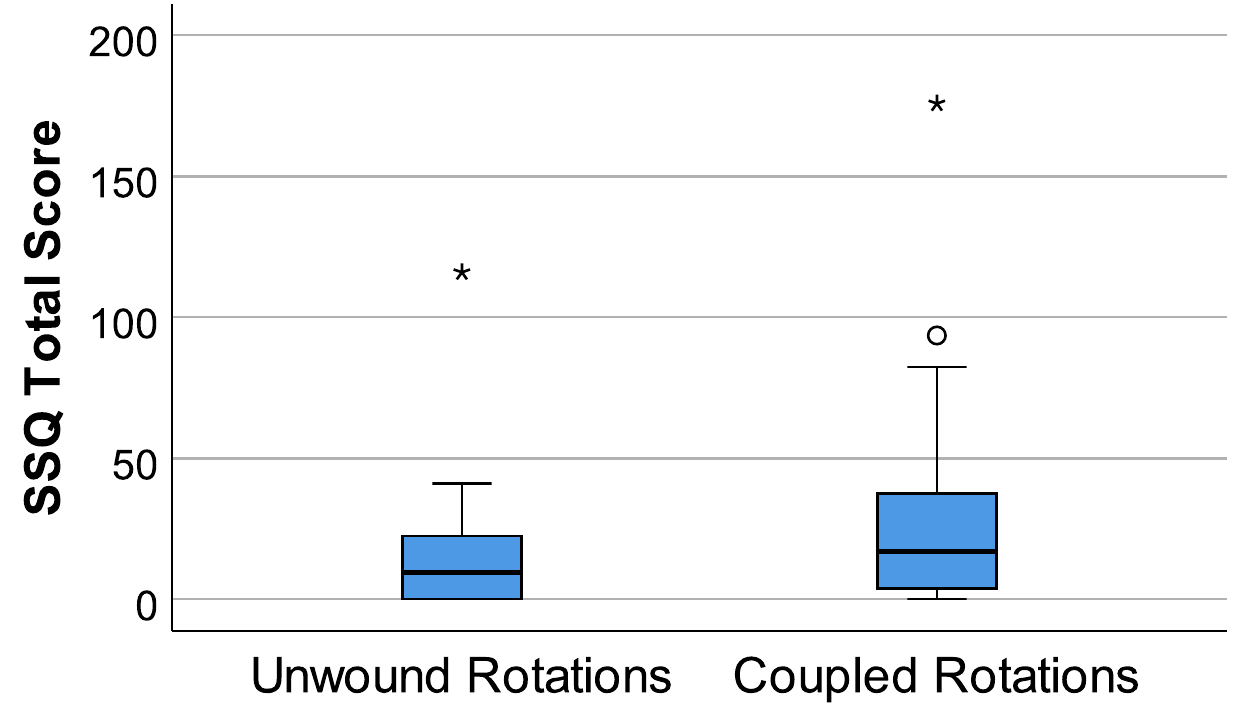}
         \caption{}
         \label{fig:SSQ_TS}
     \end{subfigure}
     \hfill
     \begin{subfigure}{0.49\columnwidth}
         \centering
         \includegraphics[trim=0 5 0 0,clip,width=\textwidth]{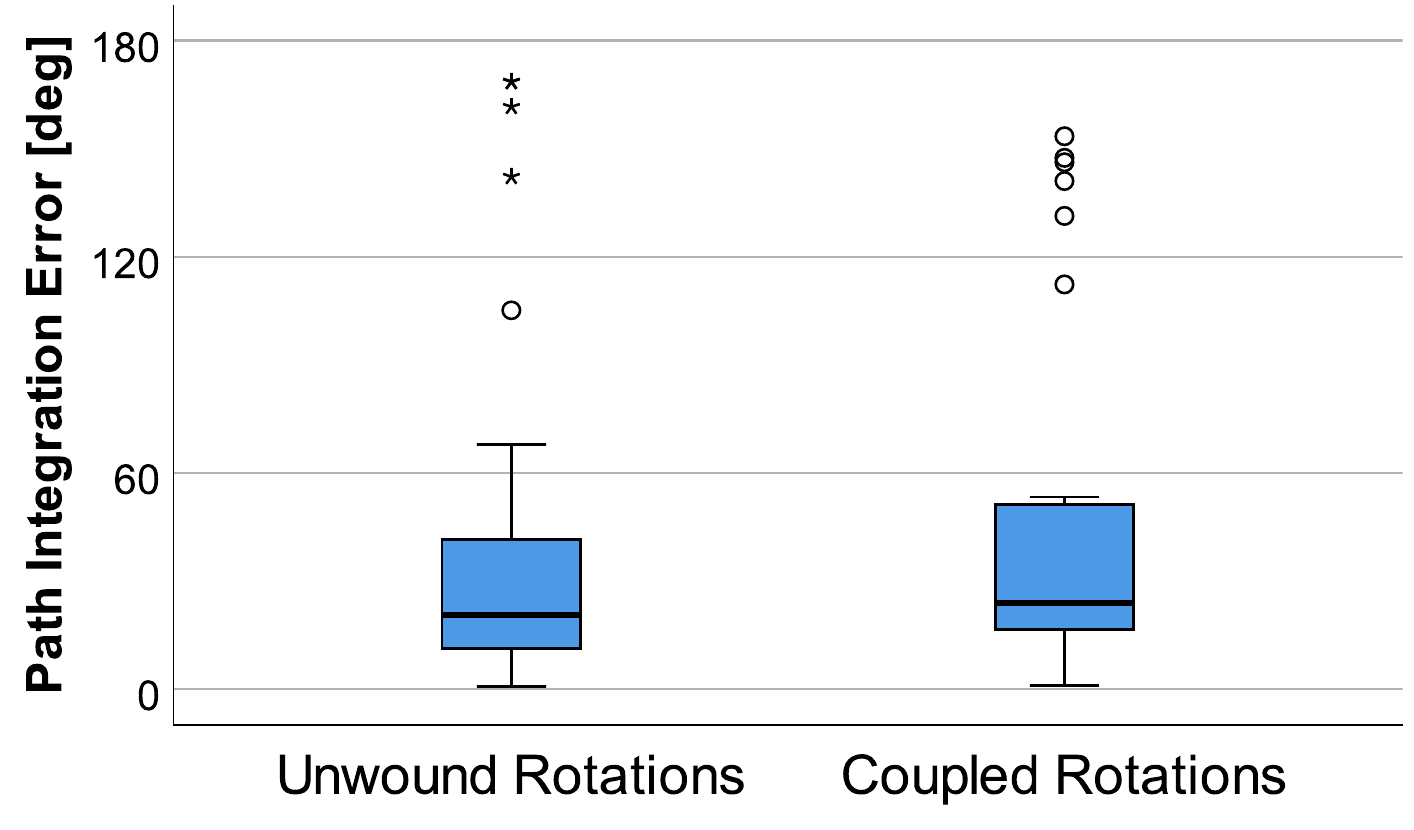}
         \caption{}
         \label{fig:path_error}
     \end{subfigure}
     \caption{Comparison of (a) total weighted SSQ scores, and (b) path integration errors.}
     \vspace{-0.5cm}
\end{figure}

A Wilcoxon Signed-Ranks test (two sided) is performed to compare the differences between the total weighted SSQ scores for UR (${Mdn}=9.35$) and CR (${Mdn}=16.83$) conditions (see Fig.~\ref{fig:SSQ_TS} for the score distributions). The test indicated that \ac{ur} elicited significantly lower SSQ scores compared to \ac{ar}, $Z=-3.46$, $p=.001$, $r=0.59$.


\begin{figure}[t!]
\centering
\includegraphics[trim=30 8 150 9,clip,width=0.9\columnwidth]{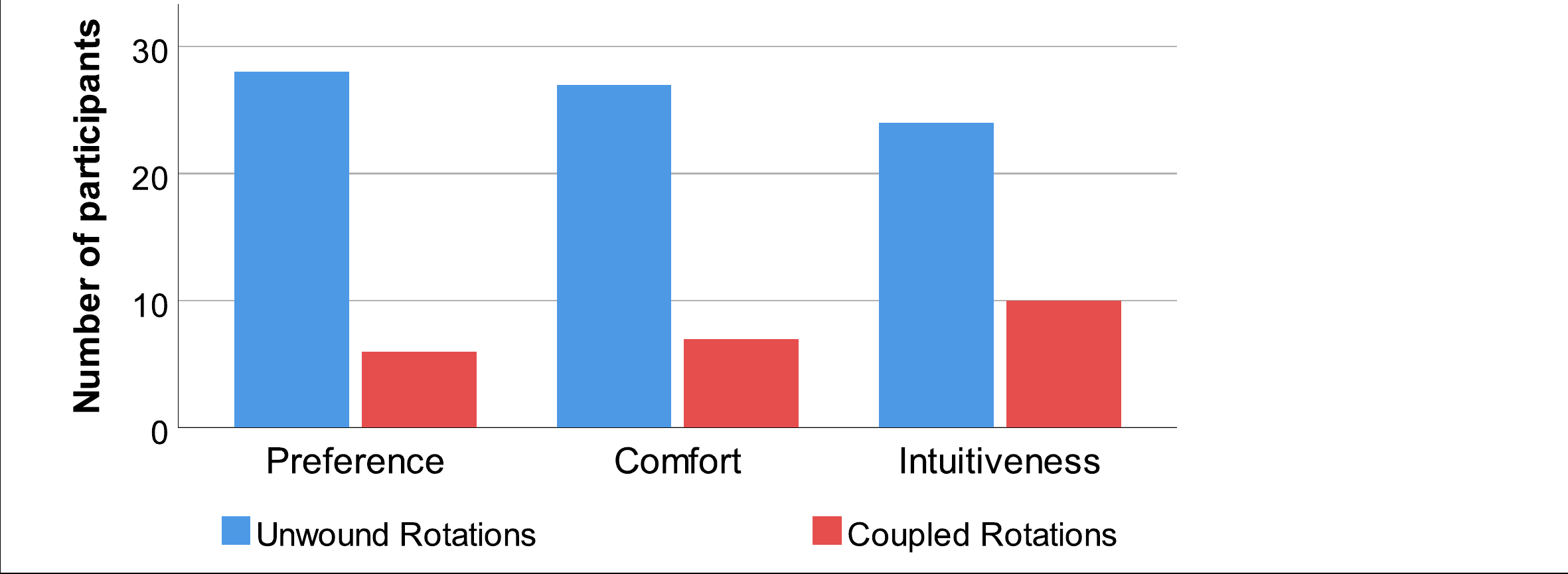}
\caption{The distributions of responses to the questions regarding preference, comfort, and intuitiveness.}
\label{fig:pref_comf}
\vspace{-0.5cm}
\end{figure}

Fig.~\ref{fig:pref_comf} shows the distributions of the responses given by the participants to the forced-choice questions regarding preference and comfort.
When asked ``Which of the two videos do you prefer?" 28 out of 34 participants $(82.4\%)$ selected the \ac{ur} condition. An exact binomial test with exact Clopper-Pearson $95\%$ CI indicated that this tendency in preference is statistically significant, $p=.000$ (two-sided, one-sided $p=.000$) and had a $95\%$ CI of $65.5\%$ to $93.2\%$. 
Consequently, 27 participants $(79.4\%)$ selected the \ac{ur} condition in response to the question ``Which of the two videos is more comfortable?". An exact binomial test with exact Clopper-Pearson $95\%$ CI was performed, showing that the condition \ac{ur} was found significantly more comfortable, $p=.001$ (two-sided, one-sided $p=.0005$) and had a $95\%$ CI of $62.1\%$ to $91.3\%$. 

The medians of the path integration error distributions for \ac{ur} and \ac{ar} conditions were $20.57^\circ$, and $23.9^\circ$, respectively (see Fig.~\ref{fig:path_error} for the error distributions).
A Wilcoxon Signed-Ranks test (two sided) indicated that \ac{ur} did not elicit any significant change in the path integration errors, $Z=-0.71$, $p=.478$, $r=0.12$. 


\subsection{Exploratory Results}\label{sec:exploratory_results}
In addition to the confirmatory analyses we performed exploratory analyses to interpret the results better.

\subsubsection{Quantitative Data}
We checked if there were any carryover effects on sickness with 5-10 minute breaks and did not observe
an order effect in the weighted total SSQ scores across conditions \ac{ur} and \ac{ar}, as indicated by a Wilcoxon Signed-Ranks test (two-sided), $Z=1.1$, $p=.272$, $r=0.19$. 
We also looked for the potential effect of gender on sickness. For this analysis, 
we used only the gender-specified participants $(n=32)$
, balanced equally between men and women. 
For \ac{ar} condition, we did not observe any significant difference between the total weighted SSQ scores of women $(Mdn=33.66)$ compared to men $(Mdn=14.96)$, as indicated by a Mann-Whitney U test (two-sided), $U=76.5$, $Z=-1.96$, $p=.051$, $r=0.35$. 
Similarly, there was not a statistically significant difference between the SSQ scores of men $(Mdn=3.74)$ and women $(Mdn=13.09)$ in \ac{ur} condition either, as indicated by a Mann-Whitney U test (two-sided), $U=98.5$, $Z=-1.14$, $p=.26$, $r=0.2$. 

In addition to the forced-choice question, Likert-scale is used to measure the user comfort. Comparing the comfort ratings in \ac{ur} condition $(Mean=5.94)$ with the ones in \ac{ar} condition $(Mean=5.54)$, we found that \ac{ur} elicited a statistically significant increase in the comfort rankings, as indicated by a Wilcoxon Signed-Ranks test (two sided), $Z=2.16$, $p=.032$, $r=0.37$. 


We observed an order effect in the path integration errors. Watching the same environment twice resulted in a statistically significant decrease in the errors after the second video compared to the first one, as indicated by a Wilcoxon Signed-Ranks test (two-sided), $Z=-2.11$, $p=.035$, $r=0.36$. However, we remind the reader that the experiment was counterbalanced such that half of the participants had $\ac{ur}$ as the starting condition and the remaining half had $\ac{ar}$ to keep potential order effects equal in both conditions. 


The head motion data is analyzed to see whether people aligned themselves with the robot heading in $\ac{ur}$ condition. 
The distributions of average deviations from the robot heading for both conditions, \ac{ur} and \ac{ar}, followed a normal distribution as indicated by a Shapiro-Wilk test, $W(34)=0.94, p=.055$, $W(34)=0.97, p=.45$, respectively. Therefore, a paired t-test was run to determine whether there was a statistically significant mean difference between the average deviations in two conditions. The mean average deviation was higher in \ac{ur} condition $(32.14^\circ \pm 13.16^\circ)$ as opposed to in \ac{ar} condition $(27.53^\circ\pm12.46^\circ )$; a statistically significant increase of $4.6^\circ$ ($95\%$ CI, $0.76^\circ$ to $8.43^\circ$), $t(33)=2.44$, $p=.02$, $d=0.42$. 


We tested whether one condition felt more intuitive for the participants by explicitly asking: Which of the two experiences feels more intuitive for you? 24 participants $(71\%)$ felt that the condition \ac{ur} felt more intuitive (Fig.~\ref{fig:pref_comf}). An exact binomial test with exact Clopper-Pearson $95\%$ CI indicated that this tendency to pick \ac{ur} condition was significant and had a $95\%$ CI of $52.5\%$ to $84.9\%$, $p=.024$ (two-sided). 

We also asked Likert-scale questions about how the users perceived the speed, distance to walls and distance to people while moving with the robot. The statistics can be seen in Table~\ref{tab:attributes}; as mentioned in Section~\ref{sec:setup}, the values for distances and speed were mainly based on \cite{mimnaugh2021analysis}. The distance to walls and speed were considered, on average, suitable (median 4 on 7-point Likert scale), whereas the distance to people was on average considered slightly too short (median 3). 

\begin{table}
\centering
\begin{tabular}{l|l|l|l|l}
                         & \textbf{Value} & \textbf{Median} & \textbf{Variance} &  \\ \cline{1-4}
\textbf{Distance to walls}  & min. 0,9~m      & 4                     &     0.78               &  \\ \cline{1-4}
\textbf{Distance to people} & min. 1~m        & 3                     &         0.86           &  \\ \cline{1-4}
\textbf{Speed}           & max. 1~m/s      & 4                     &         0.94           &  \\ \cline{1-4}
\end{tabular}
\caption{\small{The values of certain attributes of the path, and the median and variance of 7-point Likert scale subjective opinions (1 meaning too small/too close).}}
\label{tab:attributes}
\vspace{-0.4cm}
\end{table}

\begin{figure}
    \centering
    \includegraphics[trim=2 3 2 2,clip,width=0.8\columnwidth]{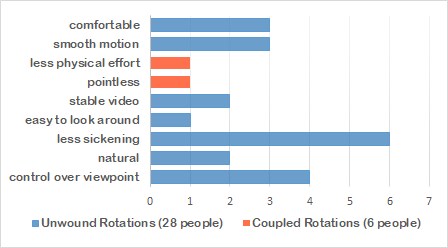}
    \caption{Frequently found codes for question ``Please explain why you prefer that video.”.}
    \label{fig:pref_oe_responses}
    \vspace{-0.55cm}
\end{figure}

\begin{figure}[t!]
    \centering
    \includegraphics[trim=2 3 2 2,clip,width=0.8\columnwidth]{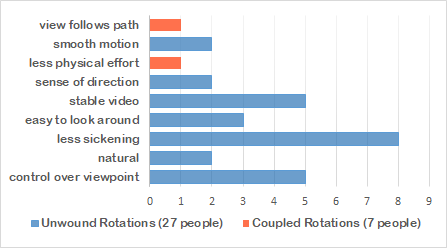}
    \caption{The frequently found codes from the question ``Please explain why that video is more comfortable.” which was asked after the participant watched both videos.}
    \label{fig:comf_oe_responses}
    \vspace{-0.55cm}
\end{figure}

\subsubsection{Qualitative data}
The open-ended data was analyzed using the thematic analysis method with inductive approach \cite{patton2005qualitative}. In the first stage of analysis, two researchers independently identified codes from the response data and mutually agreed on the codings in the second phase. It was not mandatory for participants to answer the open-ended questions, and these fields were often left blank.

The frequent codes used in open-ended question asking why participants preferred the video can be seen in Fig.~\ref{fig:pref_oe_responses}, divided by which video was preferred. The greatest number of comments (6 comments given by $21\%$ of the 28 participants who preferred \ac{ur}) said that \ac{ur}
was less sickening (for example,\textit{``Because I have to turn around by myself, so it will reduce the vertigo feeling."}). The next most frequently mentioned keywords were control over viewpoint (4 ($14\%)$,\textit{``My head orientation was not forced"}), comfort (3 ($11\%$),\textit{"I think second video is more comfortable for eyes when the scene turning"}) and smoothness of the motion (3 ($11\%$), \textit{``If I try to remember, there might have been smoother movement in the second video."}). In contrast, one person preferred CR because the physical effort of rotating in the chair felt pointless. 

The responses to open-ended questions regarding choosing one video as more comfortable are seen in Fig.~\ref{fig:comf_oe_responses}. Here, less sickening was even more evidently the biggest factor (8 ($30\%$)), with control over viewpoint being the second (5 ($19\%$)) with similar arguments as in preference. Codewords frequently not found in the question for preference but found here were stable video (5 ($19\%$),\textit{``The second one was not stable, the view kept moving, was terrible."}) and easiness to look around (3 ($11\%$),\textit{``It was easier and more natural to look around and I experienced no motion sickness."}).

\section{Discussion}
\subsection{Original hypotheses}
The results confirmed all hypotheses \textbf{H1}-\textbf{H4}. We were confident that \textbf{H1} would hold due to evidence from literature. However, whether less sickness would be enough to confirm \textbf{H2} and \textbf{H3} was an open question, since the immersion of an \ac{hmd} and the need to physically rotate to look towards the direction of motion could have caused unexpected adversarial effects. Eventually, there was only one comment in the open-ended questions about not enjoying turning in the chair (\textit{``Since I have no control on the direction, I don't want to have to turn the chair towards the direction of the video."}), but, as seen in Figs.~\ref{fig:pref_oe_responses} and \ref{fig:comf_oe_responses}, more people enjoyed being in control of the viewpoint and did not mind having to rotate the chair (\textit{``The first one felt much smoother and more comfortable. I did not mind having to turn around in my chair to look to the direction I was going.",``I preferred the lack of forced turns."}).

\begin{figure}
     \centering
     \begin{subfigure}{0.49\columnwidth}
         \centering
         \includegraphics[trim=60 75 60 60,clip,width=\textwidth]{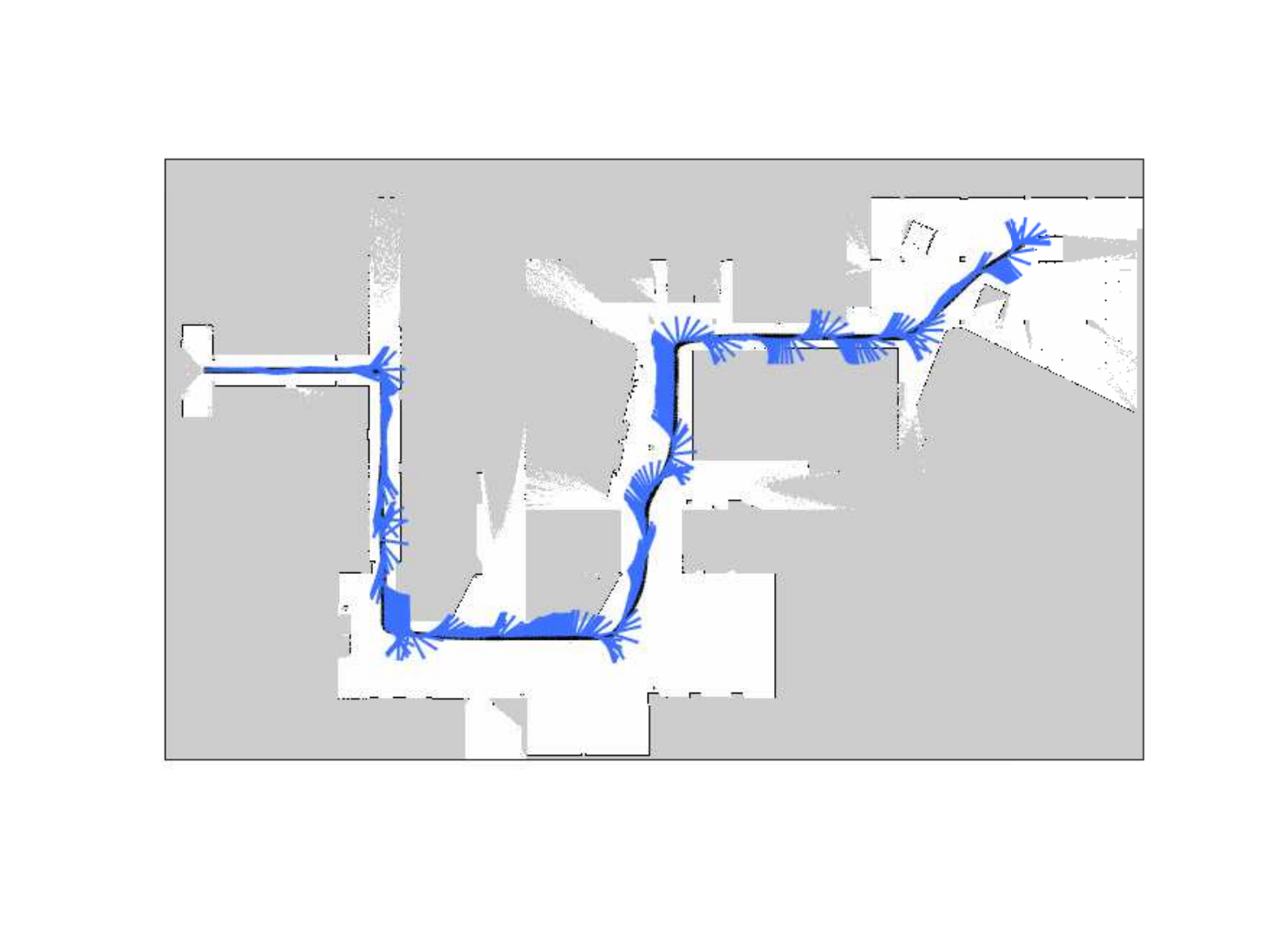}
         \caption{P1 \ac{ur}}
     \end{subfigure}
     \begin{subfigure}{0.49\columnwidth}
         \centering
         \includegraphics[trim=60 75 60 60,clip,width=\textwidth]{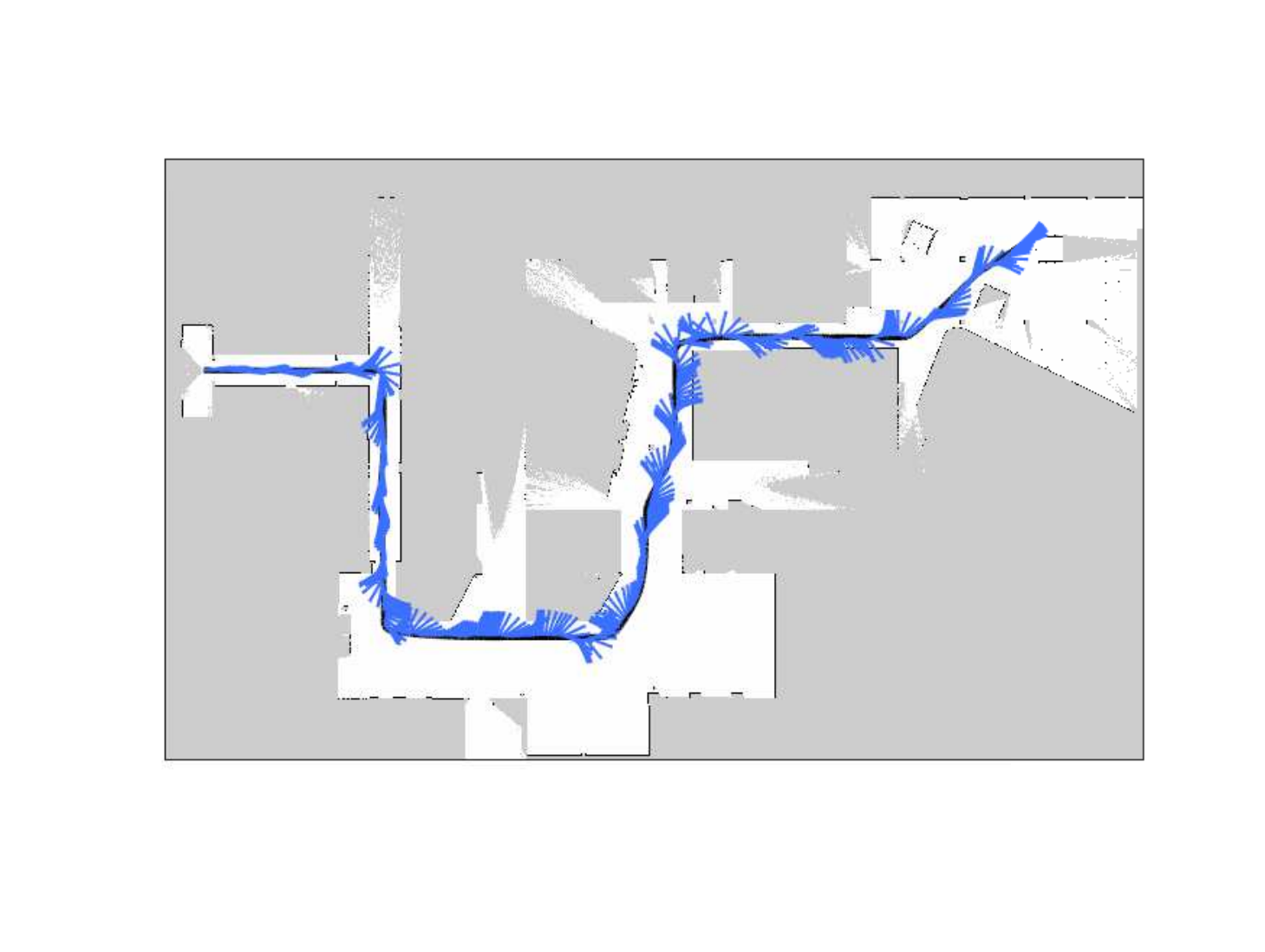}
         \caption{P1 \ac{ar}}
     \end{subfigure}
     \hfill
     \begin{subfigure}{0.49\columnwidth}
         \centering
         \includegraphics[trim=60 75 60 60,clip,width=\textwidth]{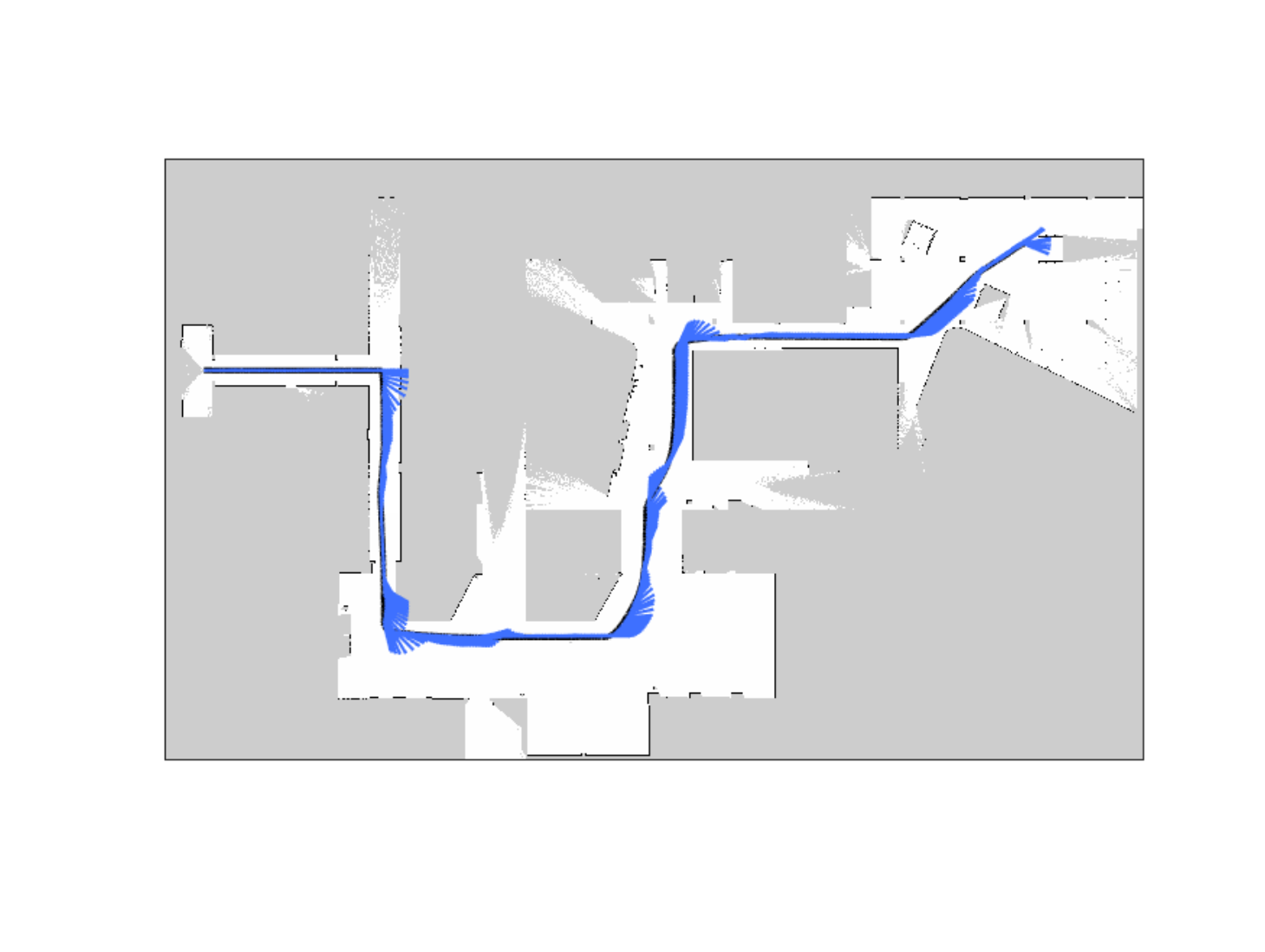}
         \caption{P2 \ac{ur}}
     \end{subfigure}
     \hfill
     \begin{subfigure}{0.49\columnwidth}
         \centering
         \includegraphics[trim=60 75 60 60,clip,width=\textwidth]{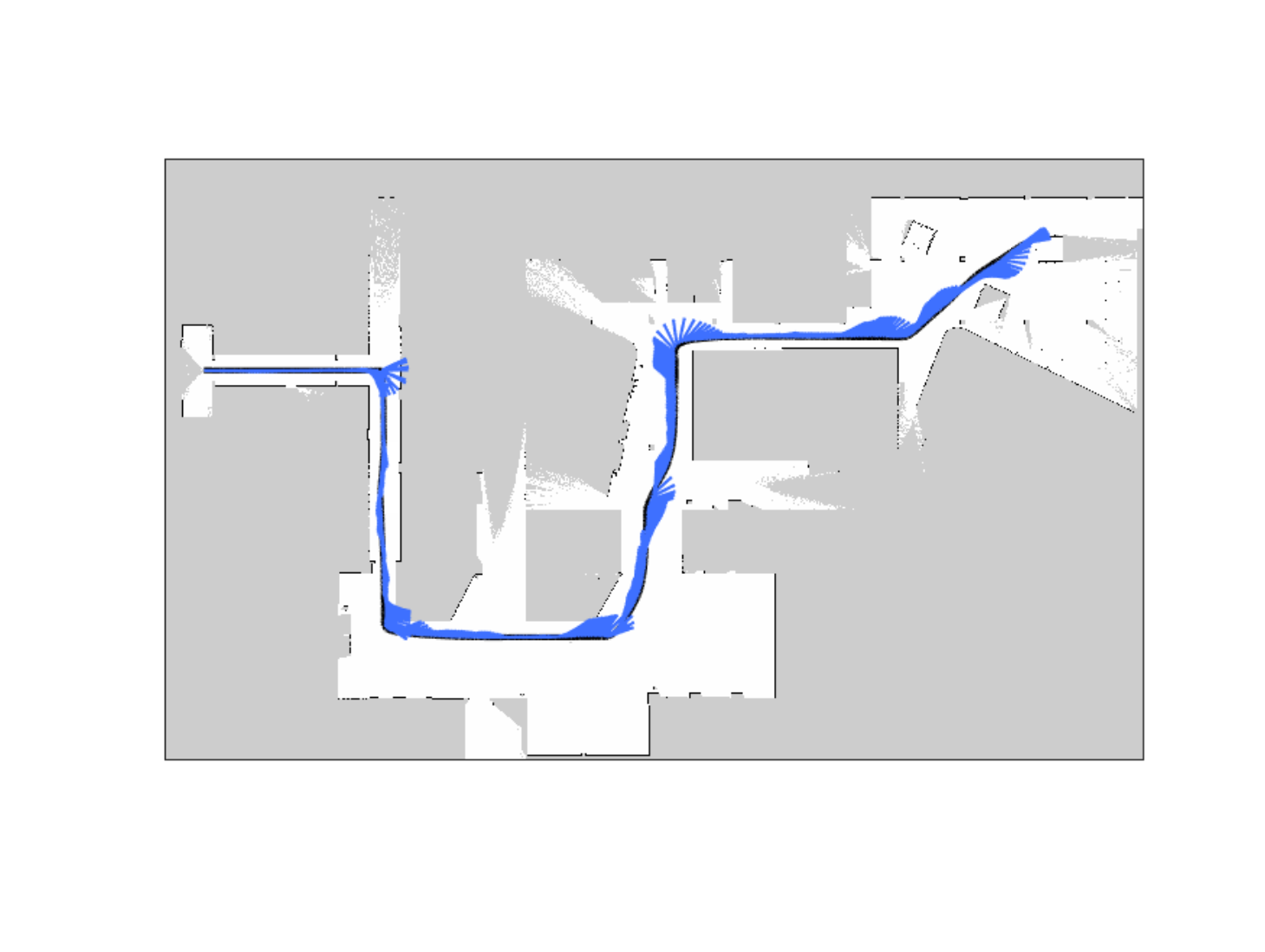}
         \caption{P2 \ac{ar}}
     \end{subfigure}
\caption{
Viewpoints of participants P1 and P2 along the robot's path.}
\label{fig:gaze_directions}
\vspace{-0.25cm}
\end{figure}

We predicted, as stated in \textbf{H4}, that 
UR does not deteriorate the spatial awareness of the users, and the results confirmed it. Even though there is some evidence in literature that physical rotations increase spatial awareness \cite{riecke2010we}, we did not explicitly tell the participants to always turn to face the direction of motion of the robot. Thus, if the participants decided not to follow up with the robot's motion, it could have made path integration more challenging. 
However, the increase in the mean of average deviation from the robot heading in UR as compared to CR was less than $10^\circ$. This is negligible considering the total amount of rotation that the robot undergoes (438\textdegree) and the corresponding compensatory motions the user needs to do. Thus, we conclude that people aligned themselves with the robot as they would if the rotations were not unwound. This is supported by the qualitative analysis of the plots that show viewpoints along the robot path (see Fig.~\ref{fig:gaze_directions}).
The users were looking forwards approximately equally
in both conditions regardless of how much they looked around, prompting us to deduce that the users rotated to face the robot's direction of motion.
Moreover, both the score on intuitiveness being in favor of UR, as well as some open-ended answers regarding intuitiveness ( \textit{``In the first video from what I can recall, I barely moved the chair. Nonetheless, on the second video I did move it every time the video changed directions, even though the instructors did not prompt me to do so.}) strengthened our belief that users do find it intuitive and natural to rotate along with the robot to help confirming \textbf{H4}. 

\subsection{Exploratory data}

We also wanted to contribute to the literature in terms of the recommended values for the clearance and speed of an immersive telepresence robot, since there is little research on the topic. 
The closest distance to walls in this study ($0.9 m$) was significantly larger than the one suggested in \cite{mimnaugh2021analysis} ($>0.4$ m) for an immersive telepresence situation. Even though there were some comments like \textit{``On some moments it felt too close. For instance, in the first turn, it felt as If I was going to crash."}, the median rating was $4$ (not too close or not too far). This suggests that $0.9$ m was an appropriate distance to keep from inanimate objects and walls. Interestingly, despite the slightly larger distance from the (simulated) people ($1\ m$), participants considered that the robot passed slightly too close $(Mdn=3)$. This is reflected also in their responses to the open-ended questions (for example, \textit{``For some of the people crossing, it felt as If I should step a bit away, as I perceived them a bit too close as well. But overall it was okay."} and \textit{"felt like I was going to run into the woman in the hallway"}). Based on the results in this study we recommend that a minimum of $1\ m$ distance should be kept from people, but further research is needed to determine an appropriate value. Whereas there is research on human-robot proxemics in VR \cite{li2019comparing}, it lacks the telepresence perspective. Finally, most of the participants rated the robot speed of $1\ m/s$ as suitable $(Mdn=4)$ with several answers to the open-ended question similar to \textit{``It felt like a walking speed of a normal person."}. On the other hand, some people perceived it to be slightly too slow (\textit{``It was slightly too slow, but not terrible."}) or too fast (\textit{``It could be a bit slower so we could see the environment with more details"}). Enabling the user to adjust the speed, or having an adaptive speed controller \cite{batmaz2020automatic} can remedy the discrepancy in the perceived speed among users (as supported by higher variance).

We did not observe carryover effects in the SSQ scores. In literature, a large variation in recovery times has been observed, with recovery times shorter than 10 minutes risking carryover effects \cite{duzmanska2018can}; in this study the breaks between videos were approximately 5-10min. One possible reason for us not observing carryover effects, regardless of the short and uncontrolled recovery time, is that the sessions were short with the 2min 18s duration; in many publications, the sessions are longer, and there are suggestions that even 55-70 minute sessions times would be acceptable to avoid excessive VR sickness
\cite{kourtesis2019validation}. Even if there were carryover effects, the counterbalancing would counter for them. 

Until recently, women were considered to suffer more from VR sickness compared to men \cite{rebenitsch2014individual}. However, a recent study proposes that the design of commercial \acp{hmd}, which are based on the interpupillary distance of men, can cause women to suffer more \cite{Peck_Sockol_Hancock_2020}. Despite observing no significant differences in total weighted SSQ scores across genders, we found that the median of scores corresponding to women were higher as compared to men in both conditions. In particular, this increase was close to being borderline significant for the \ac{ar} condition. This result was not surprising and can hopefully help designing future studies and better \acp{hmd}. 

In the open-ended answers for both comfort and preference there were six comments (the same person for both questions) justifying their choice by the order of the videos, such as \textit{"I was more familiar with the setting since I had seen it before."} and \textit{"It was more exciting because it was the first time for me to wear vr-glasses and the second one was more like repeating that experience"}. The choices for both questions were four people choosing UR and two people choosing CR; this shows that the counterbalancing worked and the order effect did not corrupt the results. Evidently, using the same environment and the same robot path to avoid confounding factors lead to directing participants' attention to aspects that were tangential to the study. Even though counterbalancing ensured that this occurred approximately evenly for both conditions, injecting small changes to the environment in between conditions can be considered for future studies.

\subsection{Limitations and Future Work}
The observed order effect in the path integration could have been avoided by telling participants before the first video exactly what would be asked from them afterwards. The participants were not told their error after the first video, but simply focusing more on the second video caused the observed order effect. Nonetheless, due to counterbalancing there is no reason to doubt the overall result of not having a significant difference in the errors. Thus, it seems that path integration mainly depends on how much the participants focus on it, and not as much on physical or automatic, coupled rotations. 

An immediate future work is to test the proposed method in real world experiments using a system comprising a wheeled robot equipped with a 360\textdegree\, camera, \change{and} an \ac{imu} to estimate the rotations that are applied to the camera frame. Although there is some evidence that 360\textdegree~videos induce more sickness than virtual environments \cite{saredakis2020factors}, the physiological causes of VR sickness remain the same. Thus, we expect unwinding rotations to improve the experience, despite the likelihood of having more vibrations even with filtering. Further future work involves applying the method proposed in this paper to robots that are capable of rotating about more axes than just one, such as a drone, \change{and researching whether the method can be applied to manually controlled vehicles as well.}


We decided not to administer the participants the Slater-Usoh-Steted (SUS) \cite{slater1994depth, usoh:2000} presence questionnaire, since we did not expect a significant difference in presence between conditions. 
However, there are some hints in the open-ended questions pointing towards increased presence with UR, such as \textit{"I felt more immersed and I didn`t feel the changes of direction as I felt in the second video."} and \textit{"It feels like I was more in control of the experience than in the second video. It felt like I was more present and could effect the experience better."} (we note that "immersion" is often understood as what we defined as "presence"). One possible reason why this could affect presence is that when users rotate more to match with the robot's direction of motion, the \textit{sensorimotor contingencies}, meaning that the user's physical actions are matched visually in the virtual world, become more utilized; for example, one user mentioned that \textit{"Because I could turn around in my chair physically, I could adjust easily where I was looking. Of course I could turn around in my chair during the second video as well, but it felt futile to do so."}. Sensorimotor contingencies have been shown to be an important building block of presence \cite{slater_psi:2009}. An interesting follow-up study would be to compare the presented approach to an automatically turning chair \cite{gugenheimer2016swivrchair}; in theory the suffered VR sickness should be the same, but we may observe a difference in presence. 

The general concept of immersive telepresence requires further research on many fronts to become competitive against current commercial flat-screen alternatives, especially for tasks 
that require
bidirectional communication. 
It is simple to show the remote user's face on a screen carried by the robot using standard solutions. However, immersion brings two issues: first, if a 360\textdegree~camera is used, should the people around the robot know where the remote user is looking at? Even though we are not aware of a research confirming this, the likely answer is yes. In this case, research should be done to reveal, for example, whether a simple solution of using an array of lights 
indicating
the remote user's view direction would suffice. Alternatively, 
a rotating screen could be attached to the robot, 
showing the remote user's face in the direction that he is looking at.
However, this brings up the second problem; when the user is wearing an \ac{hmd}, the user's face cannot be shown as simply. Fortunately, there is research towards rendering the face of an HMD-user on a screen as if she is not wearing an \ac{hmd} \cite{matsuda2021reverse}. As these technologies would likely ease two-way communicatiton with immersive telepresence robots, we believe they will gain in popularity in the near future.


\section{Conclusion}\label{sec:con}
We show that unwinding the rotations decreases VR sickness, increases user comfort, and is preferred by the users when immersed in an autonomously moving telepresence robot. Further, users physically rotate themselves to face the direction of motion even without explicit instructions to do so and find it an intuitive way to embody an autonomously moving robot. We also observed no differences in path integration abilities between unwinding rotations and letting the user's viewpoint rotate. Thus, unwinding rotations shows great potential to make immersive telepresence more popular. 

\bibliographystyle{IEEEtran}
\bibliography{references}
\end{document}